# Robust Multi-view Pedestrian Tracking Using Neural Networks

Md Zahangir Alom and Tarek M. Taha
Department of Electrical and Computer Engineering, University of Dayton, OH
Email: {alomm1,ttaha1}@udayton.edu

*Abstract*— In this paper, we present a real-time robust multi-view pedestrian detection and tracking system for video surveillance using neural networks which can be used in dynamic environments. The proposed system consists of two phases: multi-view pedestrian detection and tracking. First, pedestrian detection utilizes background subtraction to segment the foreground blob. An adaptive background subtraction method where each of the pixel of input image models as a mixture of Gaussians and uses an on-line approximation to update the model applies to extract the foreground region. The Gaussian distributions are then evaluated to determine which are most likely to result from a background process. This method produces a steady, real-time tracker in outdoor environment that consistently deals with changes of lighting condition, and long-term scene change. Second, the Tracking is performed at two phases: pedestrian classification and tracking the individual subject. A sliding window is applied on foreground binary image to select an input window which is used for selecting the input image patches from actually input frame. The neural networks is used for classification with PHOG features. Finally, a Kalman filter is applied to calculate the subsequent step for tracking that aims at finding the exact position of pedestrians in an input image. The experimental result shows that the proposed approach yields promising performance on multi-view pedestrian detection and tracking on different benchmark datasets.

## I. INTRODUCTION

Pedestrian tracking is an important task within the field of computer vision. The tracking of pedestrians in images is a challenging problem with many real-world applications in the fields of driver assistance, autonomous vehicles and visual surveillance. The development of high-powered computers, the availability of high quality and inexpensive video cameras, and the increasing demand of automated video analysis has generated a great deal of interest in human tracking algorithms. In its simplest way, tracking can be defined as a problem of estimating the trajectory of an object in the image plane as it moves around a scene. In other words, a tracker assigns consistent labels to the tracked objects in different frames of a video. Additionally, depending on the domain of tracking, a tracker can also be provided object-centric information, such as orientation, area, or shape of an object.

There are three key steps for analysis the video: identify the interesting moving objects, following of such objects from frame to frame, and determine of the object tracks for recognizing their behavior. All of the mentioned steps are very challenging due to complex backgrounds and dynamic movement of object in the input video. The key problem is the detection of the object when the object appears in the camera's field of view again and again. To resolve all of these challenges, we depend on the various information sources are contained in the input video. For instance, a single patch representing the object location in a single input frame from the video. The information under the input patch not only define the appearance of the target object but also it defines surrounding patches as well which define the appearance of the background. During the tracking of the different patches inside the input images, the different appearance can be observed for the same object with more appearance with background.

Nowadays, object tracking in the embedded platforms is of great importance with the increase of the use of mobile cameras. In most of the cases, these embedded systems are constrained with power and cannot run complex conventional object tracking algorithm on the low power CPUs available in the embedded system. The recent advent of neural processors can potentially reduce power consumption heavily while still capable to run of the powerful algorithms. These low power neuromorphic system is highly parallel which required suitable algorithm to map onto it. The significant step is to develop new algorithms based on neural networks which can run with very low power on embedded device such as mobile phone, robotics, unmanned aerial vehicle (UAV) and so on. Thus, the main contributions of this paper are:
1. This is a comprehensive approach for detection and tracking of object in video using neural networks.
2. Performance evaluation of the proposed approach on different benchmark datasets.

The rest of the paper is organized according to the following ways: In Section II, related work and background study is given in details and an explanation of the pedestrian detection system. The detailed mathematical explanation of relevant theories is given in Section III. A discussion on the dataset and experimental results is given in Section IV. Finally, conclusions and future directions are discussed in section V.

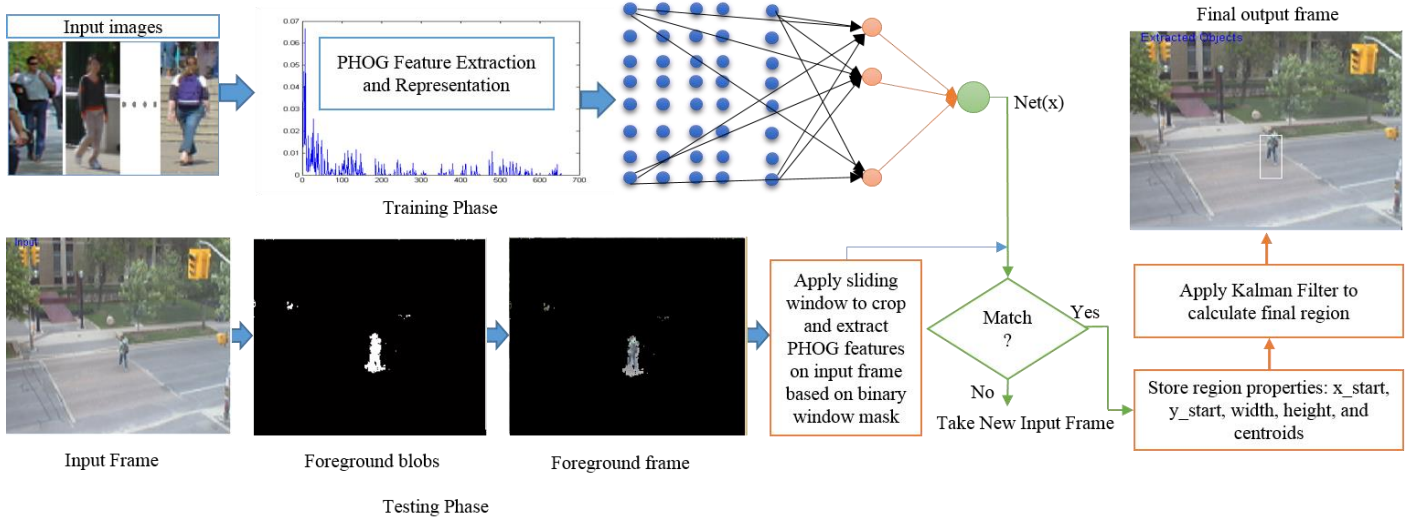

Fig. 1. Overall implementation diagram: Train Phase (a) Input image [80x32] (b) PHOG Feature representation (c) Mult-layer feed-forward fully connected Neural Network architecture. Test Phase: (a) Input Frames [320X240] (b) Foreground Blogs Extraction (c) Foreground objects

## II. RELATED WORKS

This section reviews the related approaches for each of the component of our propose system. The remaining paragraphs review the related work on pedestrian detection, Pyramid histogram of orientated gradients (PHOG), Neural Network (NN) for tracking, and Kalman filters. In the last decade, a number of pedestrian detection systems have been proposed to tackle the problem of finding humans from a moving platform which is really important for surveillance system in airport security and so on. The overviews of related work in this area can be found in [1] and [2].

Nowadays, this challenging task of finding pedestrian in the input frames received huge attention in the field of machine learning and computer vision community. There are many difficulties can be rise for pedestrian tracking such as collision avoidance, dark or fuggy input video, some of the researches are conducted to overcome those complexity for pedestrian detection and tracking [4, 5, 6]. Most of the research on this domain aim to find the boundary for the pedestrian that is significantly close to the manually labeled ground truth patches. The number of pedestrian has been counted based on the window appears on the input frame which exceeds the predetermined threshold [3].

The main challenge as well as the limitation is to extract the image features and another requirement to know the geometrical position of the object in advance. The sliding window based object detection system [35], which scan the entire images by the predefined window size with a stride. From input window, the existence of the object has been determined algorithmically. However, this type of approach is very expensive in term of computational complexity. For examples: Quarter Video Graphics Array (QVGA) frame, there are roughly 50,000 patches or window those are evaluated in every frame. For achieving the real-time performance, an efficient approach has been proposed called cascaded architecture [7]. The main and challenging part of this problem is to determine the background because background is far more frequently compare to objects, a classifier is separated into a number of stages where each enables to rejection of the background patches. As a result, the number of stages are reduced that have to be evaluated on average. There are some other techniques which are based on key point on desire objects and partial part based approaches. However, most of the up-to date models are used sliding approach for object detection task. In this work, we proposed very convenience and fast sliding windowing approach which can improve pedestrian detection technique in one step ahead.

The performance of a sliding window based method can be influenced by choosing various features and classifiers in case of pedestrian detection in particular. Some widely used features extracted from the raw image data HOG [12-13], Neural algorithms for tracking were developed based on Widrow's Adaline [8], a pioneering neural network combining neural architecture and a model-based structure suitable for tracking a single target. In case of implementing tracking systems, two problems need to be considered: prediction and correction. Predict problem: predict is the position of the object being tracked in the following frames, which is identify a region in which the probability of finding object is high. Correction problem: identify the target object in the immediate next frame within designated region. A very well-known solution for prediction is Kalman filter which recursively estimates the state of the dynamic system. Kalman filter have been adopted in different application domain for tracking including video tracking [9]. In addition, various multiple object tracking algorithms have been proposed in the literature using Kalman filters [10].

Along with the algorithm development, there are a lot of research on going in the hardware development for new technology. Most of the design emphasis on Brain inspired computing system using neural networks technique. These systems are biologically inspired and have been shown to be several orders of magnitude power efficient than conventional processors [11]. An example of such a processor available commercially is the IBM TrueNorth chip, released in November 2015 [12]. Thus, it is essential to develop algorithm which is suitable to map on modern hardware and provides almost same level of performance of traditional computing system with very low power. The next section represents on algorithms used for the learning and exploitation phases and presents with depth with the mathematical regular expression.

### III. ADPATIVE BACKGROUNDING

*A. Online mixture model*

As presented of Gaussian mixture model in [13]. For input images, "Pixel process" the values of a particular pixel over time i.e a time series of scalars for gray values or vectors for color pixel values. At any time, $t$, what is known about a particular pixel, $\{x_0, y_0\}$, is its history

$$\{X_1, X_2, \ldots X_t\} = \{I(x_0, y_0, i) : 1 \leq i \leq t\} \quad (1)$$

where $I$ is the image sequence. We chose to model the recent history of each pixels $\{X_1, X_2, \ldots X_t\}$, as a mixture of $K$ Gaussian distributions. The probability of observing the current pixels value is

$$P(X_t) = \sum_{i=1}^{K} w_{i,t} * \eta(X_t, \mu_{i,t}, \Sigma_{i,t}) \quad (2)$$

Where $K$ is the number of distributions, $w_{i,t}$ is an estimated of the weight ( the postion of the data accounted for by this Gaussian) of the $i^{th}$ Gaussain in the mixture at time $t$, $\mu_{i,t}$ and $\Sigma_{i,t}$ are the mean value and covariance matrix of the $i^{th}$ Gaussain in the mixture at time $t$ and where $\eta$ is a Gaussian probability density function

$$\eta(X_t, \mu, \Sigma) = \frac{1}{(2\pi)^{\frac{n}{2}} |\Sigma|^{\frac{1}{2}}} e^{\frac{1}{2}(X_t - \mu_t)^T \Sigma^{-1} (X_t - \mu_t)} \quad (3)$$

$K$ is the determined by the available memory and computation power. Presently, from 3 to 5 are used. Also the computational reasons, the covariance matrix is assumed to be of the form:

$$\Sigma_{k,t} = \sigma_k^2 I \quad (4)$$

It is considered that the pixel values of all the input channel of red, green and blue are independent and have the same variances. In the other cases, the assumption allows us to avoid a costly matrix inversion with the expense of some accuracy.

Therefore, the characteristics of the recent distribution of the value of the pixels in the scene can be observed with mixture of Gaussian. New pixel values considered as major component which is used to update the model. The new pixel value, $X_t$ is checked against the generated $K$ Gaussian distribution, until the match is found. A match of the pixel value is define within 2.5 standard deviation of a distribution. As we considered dynamic background situation, the threshold has little impact on overall performance for lighting and shadow images. The least probable distribution of the previous steps is replaced with a distribution with the current values as its mean value, an initial high variance and low prior weight, if none of $K$ distribution match with current pixel values. This is correspondent to the probability of this value with an exponential window on the previous values. The weight of $K$ distribution at time $t$ will be updated according to the following equation

$$w_{k,t} = (1 - \alpha) w_{k,t-1} + \alpha(M_{k,t}) \quad (5)$$

Where $\alpha$ is the learning rate and $M_{k,t}$ is 1 for the model which match and 0 for the remaining model. After this approximation, the weights are renormalized. The parameter of mean $\mu$ and variance $\sigma$ remained unchanged if the distribution remain same otherwise the mean and variance are updated as follows

$$\mu_t = (1 - \rho)\mu_{t-1} + \rho X_t \quad (6)$$

$$\sigma_t^2 = (1 - \rho)\sigma_{t-1}^2 + \rho (X_t - \mu_t)^T (X_t - \mu_t) \quad (7)$$

and

$$\rho = \alpha \, \eta(X_t \mid \mu_k, \sigma_k) \quad (8)$$

$\rho$ is the learning factor for adapting current distributions.

The most significant advantage of this background segmentation method is, it update the background model without destroying existing model.

*B. Background model estimation*

In this step, we would like to determine which of the Gaussian mixture are most likely produced by the background processes. Heuristically, we select the Gaussian distribution which have the most supporting evidence and the least variance. For determining that, the Gaussian distributions are ordered by the value of $w/\sigma$. This value increase both as a distribution gain more evidence and as the variance decrease. After that the distributions are sorted where the most likely background remain on the top and the less probable transient background distributions gravitate to the bottom[11]. Then the thresholding approach is applied to determine the first $B$ distribution for the background model, where

$$B = argmin_b \left( \sum_{k=1}^{b} w_k > T \right) \quad (9)$$

Where $T$ is a measure of the minimum portion of the data that should be accounted for by the background. If a small value for $T$ is chosen, the background model is usually unmodel otherwise a multi-modal distribution caused by a repetitive background motion(e.g. leave of tree, a flog in the wind etc.) The above method allows to identify foreground pixels in each new frame while updating the description of each pixel's process.

*C. Foreground image windowing*

After extracting the foreground image, the sliding windowing technique is applied over the binary foreground image $f_{xy}^b$. Let the window on input image as $f_{xy}^w$, is extracted sliding through the $f_{xy}^b$, window on binary foreground image $f_{xy}^b$ is represented as $f_{xy}^{wb}$ and summation of total number pixels of the binary window is $T_p$. Then desire output window $O_{xy}^w$, is extracted according to the following equation:

$$O_{xy}^w = \begin{cases} f_{xy}^w; & \sum_{x,y=1}^{w_x w_y} f_{xy}^{wb} \geq T_p * 50\% \\ avoid\ window; & otherwise \end{cases} \quad (10)$$

where $w_x$ and $w_y$ are height and width of the window respectively. Finally PHOG feature extraction technique has been applied on $I_{xy}^w$. Next section discusses on PHOG in details.

*D. PHOG Features extraction*

PHOG features descriptor is mainly inspired by two sources: (i) the image pyramid representation of Lazebnik et al. [15], and (ii) the Histogram of Gradient Orientation (HOG) of Dalal and Triggs [14].

The PHOG feature is computed by creating a pyramid histograms over the entire image and appending the histograms for each level of the pyramid into a single vector. For generating the pyramid the grid at level $l$ has $2^l$ cells along each dimension. Subsequently, level $0$ is represented by a $K$ vector corresponding to the $K$ bins of the histogram, level 1 by a $4K$-vector etc, and the PHOG descriptor of the entire image is a vector with dimensionality $K \sum_{l \in L} 4^l$. For example, for levels up to $L = 1$ and $K = 20$ bins it will be a 100-vector. In the implementation we limit the number of levels to $L = 3$ to prevent over fitting. The PHOG is normalized to sum to unity. This normalization ensures that images with more edges, for example those that are texture rich or are larger, are not weighted more strongly than others [16]. Fig. 1 (b) in Train Phase shows PHOG representation input image.

*E. Neural Network for pedestrian detection and tracking*

To apply a classifier or a ranking function to an image window, suitable features first have to be extracted from the window's raw image data. From our knowledge there is no related publication on image features which is suitable particularly for localization and pedestrian detection as a potential candidate. However, features with PHOG are used very often for pedestrian detection in the last decade. In the first layer of this neural network architecture, several small feature maps which are referred to as local receptive fields (LRF*)* moved in a satisfactory grid over the input image window. The second layer of the network contains of one or more fully connected output neurons, which also apply a transfer function to their input that show in Fig. 1(c) in training phase. Whereas in a standard multi-layer perceptron (MLP) the neurons of all layers are fully connected, the NN employs weight sharing, making the network less susceptible to over fitting. However one hidden unit of an MLP is connected to each input by a *separate* weight, whereas one LRF is applied to all image positions using the *same weights*, which help to reduce the number of adaptable parameters significantly during training.

Let $N_p$ be the number of image positions obtained by shifting $M = S_x \times S_y$ pixels at a step size of $D_x$ and pixels over input window size $M \times H$. Again, let $x_{ik} \in \mathbb{R}$ contains the value of the *k-th* pixel of the image patch at the *i*-th of $N_p$ position and let $l_{ik} \in \mathbb{R}$ contain the *k*-th weight of the *j*-th local receptive field. Then a NN with one output neuron applied to the pixel patches **x** of an input window can be written as:

$$net(x) = h\left(\sum_{i=1}^{N_p} \sum_{j=1}^{N_{lrf}} w_{i,j}\, g\left(\sum_{k=1}^{M} x_{ik} l_{jk} + \theta_j\right) + \varphi\right) \qquad (11)$$

with bias weight $\theta_j$, $\varphi \in \mathbb{R}$, output weight $w_{i,j} \in \mathbb{R}$ and transfer function g : $\mathbb{R} \to \mathbb{R}$ and h : $\mathbb{R} \to \mathbb{R}$ applied to layer one and two respectively. Transfer functions $g$ : R $\to$ R and $h$ : R $\to$ R applied to layer one and two respectively. The network architecture is illustrated Fig. 1(c) in training phase. After recognizing the pedestrian from the window of the foreground frame, location of the object is calculated for following parameters: x_start, y_start, height, width and centroid on x and y directions. The final localization of the object is calculated using Kalman filter based on preliminary position. Next section discusses detail about kalman filter.

*F. Kalman Filter for tracking*

The Kalman filter is a framework for predicting a process's state, and using measurements to correct or 'update' these predictions. Time Update: Discrete-time Kalman filters begin each iteration by predicting the process's state dynamically using a linear model. State prediction: For each time step $k$, a Kalman filter first makes a prediction $\hat{x}_k$ of the state at this time step:

$$\hat{x}_k = A\,\hat{x}_{k-1} + B * u_k \qquad (12)$$

Where $\hat{x}_{k-1}$ is a vector representing process state at time $k-1$ and A is a process transition matrix. $u_k$ is a control vector at time $k$, which accounts for the action that the object takes in response to state $\hat{x}_k$. B converts the control vector $u_k$ into state space [17]. In our model of moving objects on 2D camera images, state is a 4-dimensional vector $[x, y, dx, dy]$, where $x$ and $y$ represent the coordinates of the object's center, and $dx$ and $dy$ represent its velocity. The transition matrix is thus simply

$$A = \begin{bmatrix} 1 & 0 & 1 & 0 & 0 \\ 0 & 1 & 0 & 1 & 0 \\ 0 & 0 & 1 & 0 & 0 \\ 0 & 0 & 0 & 1 & 0 \\ 0 & 0 & 0 & 0 & 1 \end{bmatrix}$$

We chose to include $x$ and $y$ velocities in our model because they are useful features. The objects that we intend to track on camera, such as people has slowly changing velocities respect to the input frame; these velocities are easy to observe from a video stream. On the other hand, more complex features like acceleration are less useful, partly because they change suddenly and are harder to observe. The object's affect $x$ position: $u_k$ is just a scalar representing how much the object is expected to move along the $x$ axis in response to control. Converting $u_k$ into state space is very simple:

$$B = \begin{bmatrix} 1 & 0 & 0 & 0 & 0 \\ 0 & 1 & 0 & 0 & 0 \\ 0 & 0 & 1 & 0 & 0 \\ 0 & 0 & 0 & 1 & 0 \\ 0 & 0 & 0 & 0 & 1 \end{bmatrix}$$

Error Covariance Prediction: The Kalman filter concludes the time update steps by projecting estimate error covariance $P_k$ forward one time step:

$$P_k = A\,P_{k-1}A^T + Q \qquad (13)$$

Where $P_{k-1}$ is a matrix representing error covariance in the state prediction at time $k$. $Q$ is the process noise covariance. Kalman gain: First, the Kalman filter computes a Kalman gain $K_k$, which is later used to correct the state estimate $\hat{x}_k$ :

$$K_k = P_k H^T\,(H\,P_k\,H^T + R_k)^{-1} \qquad (14)$$

where H is a matrix converting state space into measurement space and R is measurement noise covariance. [17] Like $Q$, determining $R_k$ for a set of measurements is often difficult; many Kalman filter implementations statically analyze training data to determine a fixed $R$ for all future time updates. In this work $Q$ and $R$ initialized as [5x5] identity matrix. State Update: Using Kalman gain $K_k$ and measurements $z_k$ from time step k, we can update the state estimate:

$$\hat{x}_k = \hat{x}_k + K_k(z_k - H\,\hat{x}_k) \qquad (15)$$

Conventionally, the measurements $z_k$ are often derived from inputs. Note that the updated state, is still distributed by a Gaussian. Following section discusses details about database and experimental results.

## IV. RESULT AND DISCUSSION

In this study, we used multi-view NICTA Pedestrian dataset for training of this system. The system is trained with about 1000 multi-view positive and 2000 negative images. The size of images of database is 80x32. Some of the example positive multi-view pedestrian images are shown in the following figure.

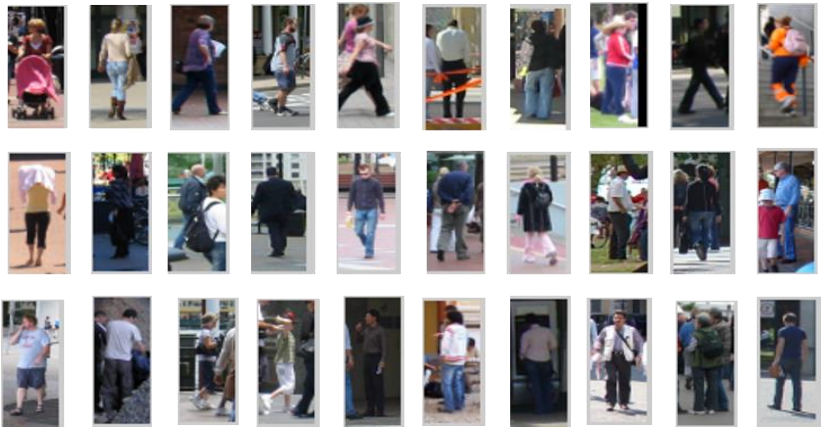

Fig.2. Example positive images for multi-view NICTA Pedestrian dataset

We trained this proposed system with multilayer neural networks consist with 2560 neurons in input layer, 200 neurons in first hidden layer, 100 neurons in second hidden layer and 2 output neurons for two classes as output. Only 50 iterations is applied to train this system with learning rate 1.

In the testing phase: we have experimented on different benchmark datasets from different sources such as TUD, PETS 2009, and ETH dataset. The system is evaluated on visual tracker benchmarks: subway, walking with huge illumination variation [19], multiple object tracking benchmarks: TUD campus sequence dataset [20], and road crossing dataset from university of oxford [21]. Some of our excremental results for tracking of mentioned dataset are shown in Fig. 3.

To evaluate the performance of our proposed system, multiple object tracking (MOT) matrices for precision and accuracy are calculated [18].
1. The multiple object tracking precision (MOTP)

$$\text{MOTP} = \frac{\sum_{i,t} d_t^i}{\sum_t c_t}$$

The total error in prediction position for matched object-hypothesis pairs over all frames, divided by the total number of matches made. It represents the ability of tracker to estimate precise object positions.

2. The multiple object tracking accuracy (MOTA):

$$\text{MOTA} = 1 - \frac{\sum_t (m_t + fp_t + mme_t)}{\sum_t g_t}$$

Where $m_t$ is for number of misses, $fp_t$ stand for false positive for time t, and $mme_t$ are the number of mismatches at time t, MOTA can be calculated according to the following three error ratios:

$$\overline{m} = \frac{\sum_t m_t}{\sum_t g_t}$$

$\overline{m}$ represents the ratio of misses in the sequence which is the total number of objects present computed over in all frames over time,

$$\overline{fp} = \frac{\sum_t fp_t}{\sum_t g_t}$$

$\overline{fp}$ is for the ratio of false positives, and

$$\overline{mme} = \frac{\sum_t mme_t}{\sum_t g_t}$$

Which is the ratio of mismatches. In the performance analysis state, MOTP and MOTA are calculated on 100 frames of each dataset. The following table shows the experimental performance in detail of this propose approach. In this implementation, the Euclidian distance measurement technique is used with mm and threshold is set to 220mm.

TABLE I. MOTP and MOTA FOR EACH DATASET

| Databases | MOTP | Miss-rate | False positive | Mismatch rate | MOTA |
|---|---|---|---|---|---|
| Road crossing | 146mm | 12.6% | 8.3% | 12.3% | 68.8% |
| Subway | 130mm | 17.6% | 13.5% | 11.7% | 57.2% |
| TUD campus | 123mm | 10.8% | 8.3% | 9.2% | 71.7% |
| Walking | 155mm | 15.2% | 14.2% | 17.3% | 53.20% |

According to the results in Table I, it is clear that highest percentage of accuracy is achieved of 71.7% in TUD campus and 68.8% in road crossing datasets respectively. Lowest performance 53.2% shows for walking dataset due to the huge variation of light sources on frames. Since the target region is selected based on the binary object mask from foreground therefore false positive rate is decreased dramatically.

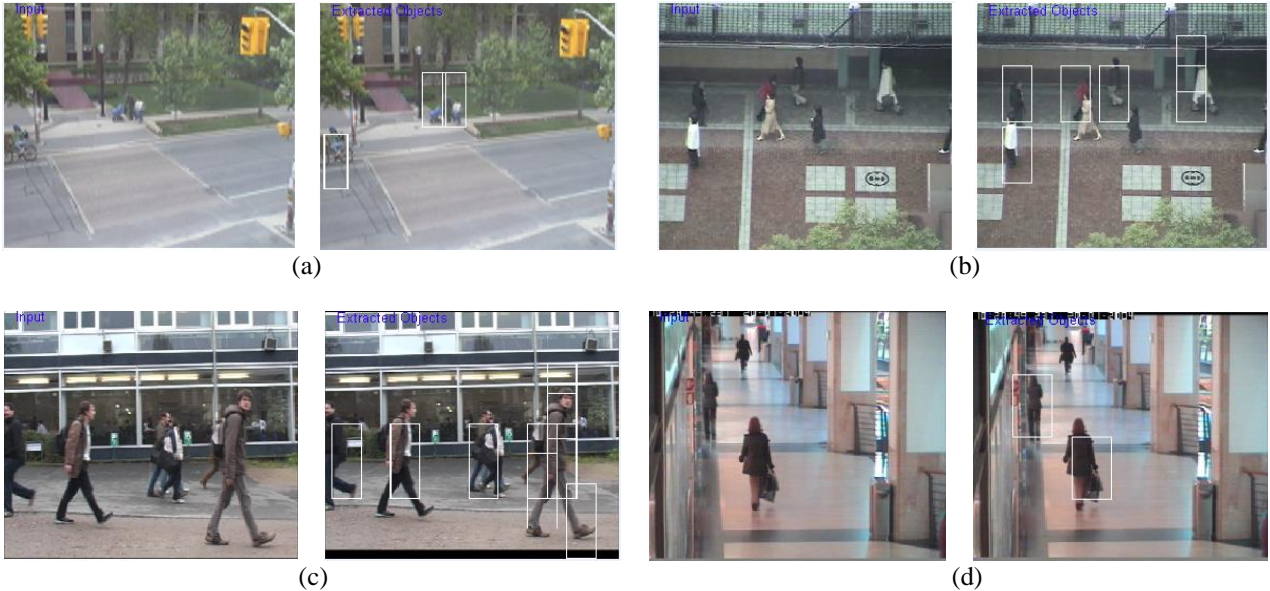

Fig.3. First column indicate inputs and second column is outputs of correspondent input: (a) Road crossing dataset, (b) Subway dataset, (c) TUD campus sequence and (d) Walking dataset.

Significant: this proposed technique is very fast which is really required for real time video surveillance system. Since we have only used the object part or partial part of the object to compute instead of all of the sliding window over the input frame. Thus the computational cost reduces drastically.

V. CONCLUSION AND FUTURE WORKS

In this paper, we proposed a robust multi-view pedestrian detection and tracking system based on foreground extraction and neural networks. Our system shows around 71.7% percentage overall tracking accuracy on TUD campus dataset. The proposed localizer chains improve significantly on the results of the multi-view pedestrian tracking. However, this proposed detection and tracking approach significantly reduces the computation cost for real-time implementation. The lackluster performance on detection and tracking on different benchmark dataset suggests that our approach needs to be further improved. One such improvement may be achieved by a combination of a richer feature extraction technique with advanced algorithm such as deep convolutional neural network (DCNN) [22]. Additionally, we will examine the implementation of this algorithm on modern neuromorphic multicore IBM's TrueNorth system [12].